\documentclass{article} % For LaTeX2e
\usepackage{iclr2026_conference,times}

\usepackage{amsmath,amsfonts,bm}

\def\eqref#1{equation~\ref{#1}}
\def\Eqref#1{Equation~\ref{#1}}
\def\1{\bm{1}}

\DeclareMathAlphabet{\mathsfit}{\encodingdefault}{\sfdefault}{m}{sl}
\SetMathAlphabet{\mathsfit}{bold}{\encodingdefault}{\sfdefault}{bx}{n}

\usepackage{amsmath}
\usepackage{amssymb}
\usepackage{mathtools}
\usepackage{amsthm}
\usepackage{multicol}
\usepackage{multirow}
\usepackage{xcolor}
\usepackage{graphicx}
\usepackage{booktabs}
\usepackage[dvipsnames]{xcolor}
\usepackage{listings}
\usepackage{upquote}
\usepackage{inconsolata}
\usepackage[table]{xcolor}
\usepackage{tcolorbox}
\tcbuselibrary{listings}
\usepackage{wrapfig}
\tcbuselibrary{listings, skins, breakable}
\usepackage{hyperref}
\usepackage[capitalize,noabbrev]{cleveref}
\usepackage{url}
\usepackage[ruled, vlined, linesnumbered]{algorithm2e}
\usepackage{xspace}
\usepackage{ulem}
\theoremstyle{remark}

\SetNlSty{textbf}{\color{blue}}{}
\SetAlgoNlRelativeSize{-1}
\SetKwSty{textbf}{\color{black}}{}
\definecolor{xxgreen}{HTML}{009F86}
\definecolor{xxpurple}{HTML}{623E99}
\definecolor{xblue}{HTML}{4169E1}
\definecolor{xsienna}{HTML}{8B4512}

\newcommand{\xxgreen}[1]{\textcolor{xxgreen}{#1}}

\newcommand{\xblue}[1]{\textcolor{xblue}{#1}}
\newcommand{\xsienna}[1]{\textcolor{xsienna}{#1}}
\newcommand{\coloredul}[2]{\textcolor{#1}{\uline{#2}}\xspace}

\lstdefinestyle{prompt}{
  basicstyle=\ttfamily\small,
  breaklines=true,
  breakindent=0pt,
  breakautoindent=false,
  frame=none,
  keywordstyle=\bfseries,
  showstringspaces=false,
  literate={~} {$\sim$}{1} {"}{{\char`\"}}{1},
}

\title{Chow–Liu Ordering for Long-Context Reasoning in Chain-of-Agents}

\author{\textbf{Naman Gupta}\thanks{Equal contribution}~~\ \ 
        \textbf{Vaibhav Singh}$^{*}$\ \
        \textbf{Arun Iyer}\ \
        \textbf{Kirankumar Shiragur}\ \
        \textbf{Pratham Grover}\\
        \textbf{Ramakrishna B.~Bairi}\ \
        \textbf{Ritabrata Maiti}\ \
        \textbf{Sankarshan Damle}\ \
        \textbf{Shachee Mishra Gupta}\\
        \textbf{Rishikesh Maurya}\ \
        \textbf{Vageesh D.~C.}\\
        \\
  Microsoft\\
}
\iclrfinalcopy % Uncomment for camera-ready version, but NOT for submission.
\begin{document}

\maketitle

\begin{abstract}
Sequential multi-agent reasoning frameworks such as \textit{Chain-of-Agents (CoA)} handle long-context queries by decomposing inputs into chunks and processing them sequentially using LLM-based worker agents that read from and update a bounded shared memory. From a probabilistic perspective, CoA aims to approximate the conditional distribution corresponding to a model capable of jointly reasoning over the entire long context. CoA achieves this through a latent-state factorization in which only bounded summaries of previously processed evidence are passed between agents. The resulting bounded-memory approximation introduces a lossy information bottleneck, making the final evidence state inherently dependent on the order in which chunks are processed.

In this work, we study the problem of chunk ordering for long-context reasoning. We use the well-known \textit{Chow--Liu trees} to learn a dependency structure that prioritizes strongly related chunks. Empirically, we show that a \textit{breadth-first} traversal of the resulting tree yields chunk orderings that reduce information loss across agents and consistently outperform both default document-chunk ordering and semantic score-based ordering in answer relevance and exact-match accuracy across three long-context benchmarks.
\end{abstract}
\section{Introduction}

Large Language Models (LLMs) \citep{yang2025qwen3technicalreport,  brown2020language, touvron2023llama, dubey2024llama, openai2024gpt4technicalreport} exhibit strong reasoning capabilities, but their performance degrades when tasks require processing context beyond the model’s (effective) input window \citep{liu2024lost,hsieh2024ruler}. Approaches such as retrieval-augmented generation (RAG) \citep{NEURIPS2020_6b493230,pmlr-v119-guu20a,ram-etal-2023-context} attempt to mitigate this issue by selecting and passing only relevant information to the model, while architectural extensions increase the amount of context that can be processed in a single pass \citep{song2024hierarchical, beltagy2020longformer,press2022train,su2024roformer}. However, these methods remain insufficient for applications where the required context exceeds even these extended limits. Sequential reasoning frameworks address this limitation by decomposing reasoning into multiple stages, allowing larger contexts to be processed through multiple coordinated LLM calls rather than a single pass \citep{wei2022chain,yao2022react,du2024improving}.

Among such sequential reasoning approaches, \textit{Chain-of-Agents (CoA)} \citep{coaneurips20204} has emerged as a representative framework for long-context reasoning. CoA partitions documents into chunks and processes them sequentially through a chain of LLM-based worker agents, each updating a shared memory state. By transforming joint reasoning over the entire context into incremental memory construction, CoA enables reasoning over inputs far exceeding an LLM’s native context window.

\begin{figure*}
    \centering
    \includegraphics[width=\textwidth]{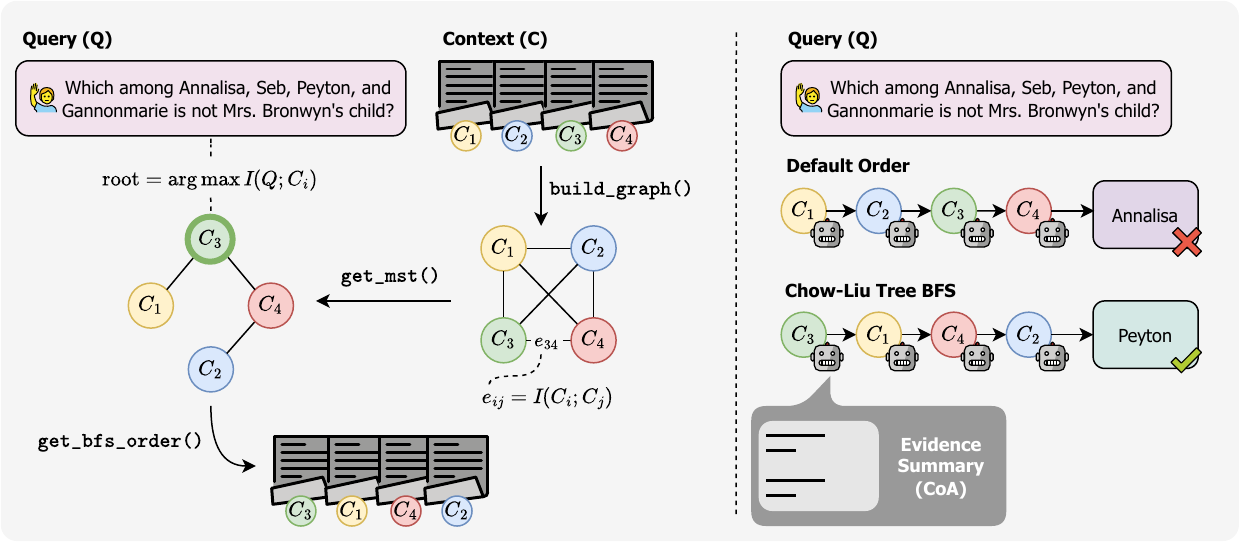}
    \caption{Overview of CoA with chunk order induced by a breadth-first traversal on the \textit{Chow--Liu} dependency tree (\textsc{CL--order}). First, we build a complete graph on chunks, where each edge models mutual information between two chunks using an embedding-based inner-product proxy. The maximum-spanning tree over the chunk graph gives the \textit{Chow--Liu} tree, which provides a second-order approximation of the global dependency structure between chunks. A breadth-first traversal on the \textit{Chow--Liu} tree rooted at chunk most similar to query yields the \textsc{CL--order}.}
    \label{fig:chow-liu}
\end{figure*}

However, this sequential design introduces a fundamental but underexplored challenge: memory is constructed through lossy and order-dependent compression. Because each update operates under a limited token budget, incorporating new information necessarily requires discarding or compressing prior content, thereby affecting how earlier context is retained and how later chunks are interpreted. Consequently, different permutations of the same chunk set can lead to distinct memory trajectories and ultimately different answers --- a sensitivity already observed in the original CoA work \citep{coaneurips20204}. Despite this, existing CoA-style approaches typically rely on default document-chunk order or naive similarity score-based orderings \citep{gupta2025cosmir}, without explicitly modeling inter-chunk dependencies.

This observation suggests that chunk ordering is not merely a preprocessing choice but a central component of reasoning under memory constraints. In this work, we take a principled approach and treat chunk ordering in CoA as a structured inference problem under a memory bottleneck. Ideally, reasoning would operate jointly over the entire set of retrieved chunks. In sequential memory construction, however, reasoning is mediated through a compressed latent memory state that must approximate the information contained in the full context. Because different chunks often depend on or complement one another, the usefulness of a chunk can depend on what has already entered memory. The ordering of chunks therefore governs how memory interacts across successive compression steps and ultimately determines how closely the final memory approximates full-context reasoning.

To address this ordering challenge, we model retrieved chunks as dependent random variables and approximate their dependency structure using a tree-structured graphical model \citep{chow1968approximating,10.1162/153244301753344605}. Specifically, we employ well-known \textit{Chow--Liu} trees \citep{chow1968approximating} to capture dominant pairwise dependencies among chunks and derive a dependency-aware processing order. Traversing this tree in a breadth-first manner from a query-relevant root keeps semantically and statistically related chunks close in the update sequence. This reduces the likelihood that complementary information becomes separated by successive compression steps. An overview of the proposed methodology is presented in Figure \ref{fig:chow-liu}, while Section \ref{sec:chowmethod} describes the approach in detail.

Empirically, we show that this dependency-aware ordering consistently improves answer relevance and exact-match (EM) accuracy across multiple long-context benchmarks \citep{zhang-etal-2024-bench,kocisky-etal-2018-narrativeqa} and model families compared to both naive document order and semantic score-based ranking. These results suggest that a significant portion of error in sequential long-context reasoning arises not only from missing relevant information but also from suboptimal ordering of interdependent chunks under memory constraints.

To summarize, our contributions are as follows:
\begin{itemize}
\item We provide a probabilistic formulation of sequential CoA-style reasoning as approximate inference over a compressed memory state, identifying chunk ordering as a key factor governing information preservation under memory constraints.
\item We introduce an efficient dependency-aware chunk ordering strategy based on a \textit{Chow--Liu} tree approximation of inter-chunk relationships.
\item We demonstrate consistent empirical gains across all evaluated models and benchmarks. In EM-based tasks, our approach outperforms default document-chunk ordering and semantic score-based ordering by 10.68\% and 6.89\% relative gains, respectively. We observe similar trends on Ragas-based benchmarks, where our method yields relative gains of 5.86\% over the default order and 6.01\% over semantic score-based baseline.
\end{itemize}

\section{Related Work}
Here we state the related work for the long-context reasoning problem.

\paragraph{Long-Context Modeling and Evaluation.}
Extending transformer models to long contexts has been widely studied. Architectural approaches include sparse attention mechanisms such as Longformer \citep{beltagy2020longformer}, improved positional encodings such as RoPE \citep{su2024roformer}, and large-scale models supporting extended windows \citep{geminiteam2024gemini15unlockingmultimodal,dubey2024llama}. However, empirical evaluations reveal that increasing nominal context length does not guarantee effective utilization. Benchmarks such as LongBench \citep{bai2023longbench}, RULER \citep{hsieh2024ruler}, and $\infty$\textsc{Bench} \citep{zhang-etal-2024-bench} show performance degradation when reasoning requires integrating information distributed across long inputs. The ``lost in the middle'' phenomenon further highlights positional biases in long-context reasoning \citep{liu2024lost}. In contrast to architectural scaling, our work addresses reasoning when total context fundamentally exceeds single-pass limits.

\paragraph{Retrieval-Augmented and Multi-Hop Reasoning.} Retrieval-augmented generation (RAG) \citep{NEURIPS2020_6b493230,pmlr-v119-guu20a,izacard2021leveraging,ram-etal-2023-context} reduces context length by selecting relevant evidence. Multi-hop retrieval systems such as Baleen \citep{khattab2021baleen} model sequential retrieval conditioned on intermediate reasoning steps. Although these methods improve evidence selection, they typically treat retrieved passages independently during ranking. Our work differs from both retrieval ranking and coherence-based ordering in focusing on
dependency-aware ordering of already-retrieved chunks under constrained shared memory, explicitly modeling statistical dependencies among chunks.

\paragraph{Sentence Ordering.} 
A distinct but related body of work studies the sentence ordering task where the goal is to recover a coherent passage from a set of unordered sentences \citep{golestani2021sentenceordering, cui-etal-2020-bert, prabhumoye-etal-2020-topological}.
These approaches focus on local semantic coherence or discourse structure in static text reconstruction settings. In contrast, our work addresses ordering under a dynamic memory bottleneck, where the sequence directly affects information preservation and reasoning outcomes in sequential LLM-based systems. 

\paragraph{Sequential and Multi-Agent Reasoning.}
Recent work decomposes reasoning into sequential or collaborative processes. ReAct \citep{yao2022react} interleaves reasoning and tool use. Multi-agent frameworks such as debate-based reasoning \citep{du2024improving} and MetaGPT \citep{hong2023metagpt} explore structured collaboration among LLM agents. Divide-and-conquer frameworks analyze when partitioned reasoning is effective under noise accumulation \citep{xu2026when}. Chain-of-Agents (CoA) \citep{coaneurips20204} provides a general long-context framework in which agents sequentially update bounded shared memory. Our work builds directly on CoA, identifying chunk ordering as a central yet underexplored design dimension in such systems.

\paragraph{Dependency Modeling and Tree-Structured Approximations.}
\textit{Chow--Liu} trees \citep{chow1968approximating} provide the optimal tree-structured approximation of a joint distribution based on pairwise mutual information. Extensions such as mixtures of trees \citep{10.1162/153244301753344605} and improved estimators for large alphabets \citep{han2023beyond} have expanded their applicability. Cutset networks further generalize tree-structured approximations for tractable probabilistic inference \citep{rahman2014cutset}. 

In this work, we adopt this principled probabilistic tool to model inter-chunk dependencies and derive ordering policies for long-context sequential reasoning systems.

\section{Problem Formulation}

Given retrieved chunks $x_{1:N} = (x_1, \dots, x_N)$ relevant to a query
$q$, the ideal answer obtainable from the full context is defined as a
sample from the conditional distribution
\begin{equation}
\label{eq:ideal}
P(a \mid q, x_{1:N}),
\end{equation}
which corresponds to a model capable of jointly reasoning over all available
chunks. In practice, however, large language models (LLMs) cannot process
arbitrarily long contexts due to finite context-length constraints.
Approaches such as Chain-of-Agents (CoA) \citep{coaneurips20204} address this limitation by
partitioning long contexts into smaller segments processed sequentially,
thereby recasting long-context reasoning as incremental memory construction.

We model this process through an \emph{idealized latent-state
factorization}, where previously processed content is compressed into a
memory state that preserves essential information while allowing new inputs
to be incorporated. Summaries provide one example of information stored in
such memory. Because memory is constructed incrementally, early compression
decisions influence how later information is interpreted and retained.
Consequently, the order in which chunks are processed directly affects the
final memory state and the resulting answer. The goal is therefore to find an
ordering that best approximates the ideal full-context inference in~\Eqref{eq:ideal} under context-length constraints.

\subsection{Sequential Memory Construction and Order Discovery}

Given an ordering $\pi$ of retrieved chunks, memory evolves sequentially as
\[
M_i^\pi = F(M_{i-1}^\pi, x_{\pi(i)}),
\]
where $F$ is an LLM-based memory update operator and $M_0$ is an initial
empty memory state. Since the memory state is represented using a token-bounded summary, F necessarily performs lossy compression of previously observed evidence. After all chunks are processed, an answer is generated
as
\[
a^\pi = G(M_N^\pi, q),
\]
where $G$ denotes the answer-generation operator, which performs reasoning conditioned only on the compressed memory state rather than the full set of chunks. Consequently, any information from the context that is not preserved in the final memory state cannot influence the generated answer, making answer quality directly dependent on how well the sequential memory construction process retains task-relevant evidence.

We assume the following sufficiency properties.

\paragraph{Incremental sufficiency (Markov property).}
\[
P(M_{i+1}^\pi \mid x_{\pi(1:i+1)}, M_i^\pi)
= P(M_{i+1}^\pi \mid M_i^\pi, x_{\pi(i+1)}),
\]
meaning that the current memory state contains all information required for
future updates.

\paragraph{Answer sufficiency.}
\[
P(a^\pi \mid x_{\pi(1:N)}, q, M_N^\pi)
= P(a^\pi \mid M_N^\pi, q),
\]
meaning that the final memory state contains all information necessary to
produce the answer.

The update operator $F$ is order-sensitive, since compression decisions made early in the sequence constrain how later evidence is integrated and retained. Different orderings lead to
different memory states and therefore different answers \citep{coaneurips20204}. Producing the ideal
answer thus critically depends on chunk ordering, making order discovery a
central algorithmic challenge.\footnote{These assumptions are modeling
abstractions rather than guarantees of actual LLM behavior; they serve only
to motivate the CoA framework and analyze how information may be preserved or
lost.}

Let $Y^\star$ denote the random variable corresponding to the ideal
full-context answer and $A^\pi$ the random variable corresponding to the
answer generated under ordering $\pi$. The objective of ordering is to ensure
that the distribution of answers produced through incremental memory
construction matches, as closely as possible, the distribution of ideal
full-context answers. In this sense, order discovery seeks an ordering whose
incremental incorporation of chunks into memory preserves the reasoning
behavior of full-context inference despite context-length constraints, which requires processing mutually dependent evidence in a manner that reduces the risk of premature compression.

\section{Embedding-Based Tree Construction for Order Discovery}
\label{sec:chowmethod}
As discussed in the earlier sections, the effectiveness of incremental memory construction depends on how well the processing order aligns with latent dependencies among retrieved chunks. Some chunks provide complementary evidence, while others only become meaningful after related information has already entered memory. A reasoning system with unlimited context can exploit these dependencies jointly, but incremental processing compresses information into a linear sequence, potentially separating related evidence across distant updates and reducing reasoning quality.

We model retrieved chunks as random variables
$X_1, \dots, X_N$ drawn from a query-conditioned distribution
$P(X_{1:N} \mid q)$,
which captures co-occurrence patterns of evidence across similar queries. Dependencies among chunks can be quantified using mutual information, where large values of \(I(X_i; X_j)\) indicate strong statistical or semantic dependence. Recovering these dependencies in principle corresponds to learning a directed acyclic graph (DAG) over \((X_1, \dots, X_N)\) that best explains their joint distribution via maximum-likelihood structure learning. However, learning general DAGs is computationally intractable.

To obtain a tractable approximation, we restrict dependencies to tree structures, inspired by the Chow--Liu algorithm. This algorithm efficiently computes the best tree-structured approximation to a joint distribution. Let \(P_T\) denote a tree-structured distribution approximating the true distribution \(P(X_{1:N} \mid q)\). Chow--Liu shows that the tree minimizing the Kullback–Leibler divergence,
\[
D_{\mathrm{KL}}\!\left(P \;\|\; P_T\right)
\]
over all trees is exactly the tree maximizing total pairwise mutual information:
\[
T^\star = \arg\max_{T \text{ tree}}
\sum_{(i,j)\in T} I(X_i; X_j).
\]
Thus, the optimal tree can be recovered efficiently using a maximum-weight spanning tree algorithm, avoiding the intractability of general DAG learning.

In practice, especially in long-context reasoning, reliable mutual-information estimates are often unavailable. We therefore use embedding similarity as a scalable proxy for mutual information. Let \(\phi(\cdot)\) denote an embedding encoder producing representations
$e_i = \phi(x_i)$.
The similarity between chunks is measured using cosine similarity,
\begin{equation}
s_{ij} =
\frac{e_i^\top e_j}{\|e_i\| \, \|e_j\|},
\end{equation}
which approximates semantic relatedness. We construct a complete weighted graph over chunk indices $\{1,\dots,N\}$ with weights $s_{ij}$ and compute a maximum-weight spanning tree
\[
T^{\mathrm{sim}} =
\arg\max_{T \text{ tree}}
\sum_{(i,j)\in E(T)} s_{ij}.
\]

This tree captures dominant relationships among chunks while remaining computationally efficient. Figure \ref{fig:chow-liu} illustrates the order discovery process using the Chow--Liu algorithm. Since the final answer is unknown at inference time, we select a traversal root based on the query. Let
$e_q = \phi(q)$
denote the query embedding, and choose
\[
r(q) = \arg\max_i \mathrm{cos}(e_q, e_i).
\]

Orienting the tree at \(r(q)\) and performing a breadth-first traversal yields a structure-aware ordering for sequential memory construction. Processing chunks in this order keeps related evidence close in memory updates, mitigating compression-induced information loss and improving answer relevance under context-length constraints. Algorithm~\ref{alg:main} outlines our proposed methodology.

\vspace{1em}
\begin{tcolorbox}[
  colback=gray!20, colframe=gray!20,
  left=5mm, right=2mm, top=2mm, bottom=1mm,
  boxsep=0pt, width=1\linewidth, before skip=2pt, after skip=2pt, 
  enlarge left by=0mm, enlarge right by=0mm,
]
  % ---- two minipages side by side ----
  \begin{minipage}[t]{0.55\linewidth}
    \begin{algorithm}[H]
    \label{alg:main}
\small
\SetAlgoNlRelativeSize{-3}
\SetNlSty{tiny}{}{}
\SetNlSkip{0.3em}
\LinesNumbered
\DontPrintSemicolon
      \caption{Chow--Liu CoA}
      \label{alg:coa-sim-tree}
\KwIn{%
  \parbox[t]{0.76\linewidth}{\raggedright
    $q$: input query; $\{x_i\}_{i=1}^N$: document chunks; $\phi$: embedding encoder.
  }%
}

\KwOut{$\hat{y}$: final answer}

Let $e_i \leftarrow \phi(x_i)$ for all $i \in [N]$ and $s_{ij} \leftarrow \mathrm{cos}(e_i, e_j)$ for all $i,j \in [N]$\;

$T^{\mathrm{sim}} \leftarrow \text{MWST}(\{s_{ij}\})$\;

$e_q \leftarrow \phi(q)$\;
$r(q) \leftarrow \arg\max_i \mathrm{cos}(e_q, e_i)$\;

$\pi \leftarrow \text{BFS}(T^{\mathrm{sim}}, r(q))$\;

$M_0 \leftarrow \varnothing$\;

\For{$k = 1$ \KwTo $N$}{
    $j \leftarrow \pi(k)$\;
    $M_k \leftarrow \mathrm{Worker}(q, x_j, M_{k-1})$\;
}

$a^\pi \leftarrow \mathrm{Manager}(q, M_N)$\;

\Return $a^\pi$\;
    \end{algorithm}
  \end{minipage}%
  \hspace{-1em}%
  \begin{minipage}[t]{0.45\linewidth}
    \vspace{0.3em}
    \small
    \begin{itemize}
        \item[] \textbf{Algorithm Overview}
        \item \xsienna{\emph{\textbf{Step: 1}}} Encode each chunk $x_i$ into embedding $e_i$ and compute pairwise cosine similarities $s_{ij}$.
        \item \xsienna{\emph{\textbf{Step: 2}}} Build a \textbf{\xxgreen{maximum-weight spanning tree}} $T^{\mathrm{sim}}$, capturing the embedding-based dependency.
        \item \xsienna{\emph{\textbf{Step: 3--5}}} Select root $r$ as the chunk most similar to $e_q$, then perform \textbf{\xblue{BFS traversal}} to obtain ordering $\pi$.
        \item \xsienna{\emph{\textbf{Step: 6--9}}} Process chunks sequentially via \textbf{CoA message passing}, where each \coloredul{xxpurple}{Worker} compresses evidence into message $M_k$.
        \item \xsienna{\emph{\textbf{Step: 10}}} The \coloredul{xxpurple}{Manager} aggregates the final message $M_N$ to produce answer $a^\pi$.
    \end{itemize}
  \end{minipage}
\end{tcolorbox}
\section{Experimental Setup}
Here we provide details of the experimental setup.

\textbf{Datasets.} 
We evaluate our approach on the long-document question answering (LongQA) benchmark HELMET \citep{yen2025helmetevaluatelongcontextlanguage}. LongQA includes English book QA and multiple choice (MC) subsets from $\infty$\textsc{Bench} \citep{zhang-etal-2024-bench} and NarrativeQA \citep{kocisky-etal-2018-narrativeqa}. In NarrativeQA, we restrict evaluation to queries with context length over $256$K tokens. These datasets feature extremely long contexts and dispersed evidence, creating a strong memory bottleneck under sequential processing, making them suitable testbeds for evaluating order sensitivity in CoA-style systems.

\textbf{Evaluation.} 
For LongQA and NarrativeQA, we use the \textit{answer relevance} metric from Ragas (Retrieval Augmented Generation Assessment) \citep{es2025ragasautomatedevaluationretrieval} --- an open-source framework designed to evaluate, benchmark LLM reasoning applications. It uses LLM-as-a-judge to measure performance metrics like answer relevance, allowing for scalable, and reliable assessment. We use Ragas instead of $n$-gram matching metrics like ROUGE F1 \citep{lin-2004-rouge} as they penalize long and paraphrased, although correct LLM generations, making it unreliable for free-form answer evaluation. In LongQA MC, we use Exact Match (EM) accuracy as the evaluation metric.

\textbf{Model Pool.} 
We use a heterogeneous model pool comprising proprietary and open-weight LLMs. Specifically, we evaluate across three models: \textsc{GPT-4.1}~\citep{openai2024gpt41}, \textsc{GPT-4.1-mini}~\citep{openai2024gpt41}, and \textsc{Qwen-3-14B}~\citep{yang2025qwen3technicalreport}. They span different capacity regimes. For computing chunk similarities, we generate dense embeddings of raw chunks using \textsc{Text-Embedding-3-Large}~\citep{openai2024textembedding3large}.

\textbf{CoA Design.} In the CoA setup, we split query-specific long-documents into chunks of 8K tokens, and process one chunk per agent step until all chunks are exhausted. Further, we allocate a memory budget of 8K tokens per agent step. For the \textsc{Qwen-3-14B} model, we limit the number of thinking tokens to 4K tokens. Additionally, we use the same prompts as proposed in CoA. To ensure reproducibility, we provide exact prompts, and hyperparameters details in Appendix \ref{app:prompt} and Appendix \ref{app:hparam}, respectively.

\textbf{Baselines.}
To assess the efficacy of Chow--Liu ordering (\textsc{CL--order}), we compare CoA with our principled approach against two other natural ordering variants: 1) \textsc{Default} and 2) \textsc{Dense}. \textsc{Default} uses CoA with the natural document-chunk order. \textsc{Dense} uses semantic score-based chunk order, where document chunks are ranked by similarity to the query.
\section{Results and Discussion}
\begin{table*}[t]
\centering
\resizebox{\textwidth}{!}{%
 \begin{tabular}{llccccccccc}
 \toprule
 & & \multicolumn{3}{c}{\textbf{\textsc{LongQA (Ragas $\uparrow$)}}} 
   & \multicolumn{3}{c}{\textbf{\textsc{LongQA--MC (EM $\uparrow$)}}} 
   & \multicolumn{3}{c}{\textbf{\textsc{NarrativeQA (Ragas $\uparrow$)}}} \\
 \cmidrule(lr){3-5} \cmidrule(lr){6-8} \cmidrule(lr){9-11}
 \textbf{\textsc{Model}} & \textbf{\textsc{Memory}} 
 & \textbf{\textsc{Default}} & \textbf{\textsc{Dense}} & \textbf{\textsc{CL--order}}
 & \textbf{\textsc{Default}} & \textbf{\textsc{Dense}} & \textbf{\textsc{CL--order}}
 & \textbf{\textsc{Default}} & \textbf{\textsc{Dense}} & \textbf{\textsc{CL--order}} \\
 \midrule
% final quant. figures retaining ids that were excluded due to cosmir
 \textsc{Qwen-3-14B}
  & CoA    
  & 41.43 & 42.25 & \textbf{44.12} 
  & 24.89 & 26.20 & \textbf{30.26} 
  & 35.72 & 38.26 & \textbf{41.23} \\
 \midrule

 \textsc{GPT-4.1-mini}
  & CoA   
  & 51.94 & 47.96 & \textbf{54.35} 
  & 65.22 & 67.39 & \textbf{70.29} 
  & 50.16 & 50.81 & \textbf{52.39} \\
 \midrule

 \textsc{GPT-4.1}
  & CoA    
  & 59.03 & 58.56 & \textbf{60.68} 
  & 82.84 & 84.33 & \textbf{85.07} 
  & 57.30 & 55.93 & \textbf{58.08} \\
 \bottomrule
 \end{tabular}
 }
 \caption{Performance on long-context benchmarks with \textsc{GPT-4.1} series and \textsc{Qwen-3}. Here, $\uparrow$ indicates that higher is better. For LongQA and NarrativeQA, we use \textit{answer relevance} metric from Ragas \citep{es2025ragasautomatedevaluationretrieval}.}
 \label{tab:main_results}
\end{table*}

\begin{table*}[t]
\centering
\resizebox{0.7\textwidth}{!}{%
\begin{tabular}{llcccccc}
\toprule
 & & \multicolumn{3}{c}{\textbf{\textsc{BM25}}} 
   & \multicolumn{3}{c}{\textbf{\textsc{Qwen-3 Embedding}}} \\
\cmidrule(lr){3-5} \cmidrule(lr){6-8}

\textbf{\textsc{Model}} & \textbf{\textsc{Memory}} 
& \textbf{\textsc{Default}} & \textbf{\textsc{Dense}} & \textbf{\textsc{CL--order}} 
& \textbf{\textsc{Default}} & \textbf{\textsc{Dense}} & \textbf{\textsc{CL--order}} \\
\midrule

\textsc{Qwen-3-14B}
& CoA
& 24.89 & 26.64 & \textbf{27.51}
& 24.89 & 26.20 & \textbf{28.82} \\

\midrule

\textsc{GPT-4.1-mini}
& CoA
& 72.83 & \textbf{75.00} & 73.90
& 73.26 & 75.58 & \textbf{77.91} \\

\midrule

\textsc{GPT-4.1}
& CoA
& 82.47 & 83.51 & \textbf{84.54}
& 82.17 & 83.16 & \textbf{85.14} \\

\bottomrule
\end{tabular}
}
\caption{Scoring function and embedding ablation on \textsc{LongQA--MC}. }
\label{tab:embedding_ablation}
\end{table*}

In this section, we present our main results. Table~\ref{tab:main_results} compares CoA with our chunk order induced by breadth-first traversal on the Chow--Liu dependency tree \textsc{(CL--order)} against two other ordering baselines: (i) default document-chunk order \textsc{(Default)} and (ii) semantic score-based dense ranking baseline \textsc{(Dense)}. We benchmark performance on \textbf{\textsc{LongQA}}, \textbf{\textsc{LongQA--MC}}, and \textbf{\textsc{NarrativeQA}} across the three LLM backbones. Significantly, across all configurations, \textsc{CL--order} consistently outperforms \textsc{Default} and \textsc{Dense}.

For instance, on \textbf{\textsc{LongQA}}, \textsc{CL--order} improves over \textsc{Default} by $+2.69$ in the \textit{Answer Relevance} metric from Ragas~\citep{es2025ragasautomatedevaluationretrieval} for \textsc{Qwen-3}, $+2.41$ for \textsc{GPT-4.1-mini}, and +1.65 for \textsc{GPT-4.1}. In contrast, \textsc{Dense} shows inconsistent gains over \textsc{Default}, with $+0.82$ points improvement for \textsc{Qwen-3}, and $-3.98$, and $-0.47$ for \textsc{GPT-4.1-mini} and \textsc{GPT-4.1}, respectively. On \textbf{\textsc{LongQA--MC}}, \textsc{Dense} improves over \textsc{Default} by $+1.31$, $+2.17$, and $+1.49$ EM points across the three models and \textsc{CL--order} further provides gains over \textsc{Dense} with $+4.06$ points in \textsc{Qwen-3}, $+2.9$ points in \textsc{GPT-4.1-mini}, and $+0.74$ EM points in \textsc{GPT-4.1}. On \textbf{NarrativeQA}, \textsc{CL--order} again showcases consistent gains over \textsc{Dense} with $+2.97$ \textit{answer relevance} in \textsc{Qwen-3}, $+1.58$ in \textsc{GPT-4.1-mini}, and $+2.15$ for \textsc{GPT-4.1}.

Overall, these results show that modeling global chunk dependencies with Chow--Liu trees and deriving chunk ordering based on this tree structure yields consistent gains over naive document-chunk ordering and ranking chunks solely on their isolated similarity score to the query. In the remainder of this section, we present ablation studies using different embedding strategies and different traversal strategies over the mutual information graph.

\begin{wrapfigure}{r}
{0.42\textwidth}
    \vspace{-8pt} 
    \centering
    \includegraphics[width=0.42\textwidth]{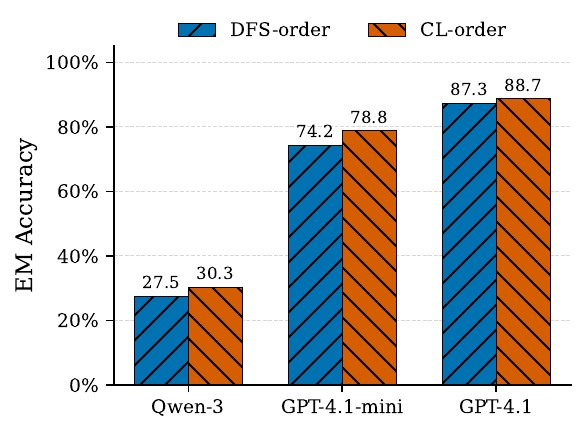}
    \caption{Comparison of DFS on the complete chunk graph against BFS on the Chow--Liu tree. EM reported on LongQA--MC for the three models.}
    \label{fig:dfsvsclplot}
    \vspace{-10pt}
\end{wrapfigure}

\textbf{Representation Ablation.} To ascertain that our findings are not an artifact of a specific dense representations from \textsc{Text-Embedding-3-Large}, we evaluate our experimental setup using (1) a sparse lexical function, \textsc{BM25} and (2) \textsc{Qwen-3-Embedding-8B} \citep{qwen3embedding}, an open-weight dense embedding model. Particularly, \textsc{BM25} replaces the semantic embeddings and embedding-based similarity scoring with a term-based ranking score between chunks. Due to the asymmetric nature of BM25 function, we compute a symmetric similarity score via the mean of the bidirectional rankings. Results on \textbf{\textsc{LongQA--MC}} are summarized in Table \ref{tab:embedding_ablation}.

With \textsc{BM25}, the efficacy of \textsc{CL--order} is not consistent against the local \textsc{Dense} ranking baseline, with $-1.1$ EM accuracy drop on \textsc{LongQA--MC} with \textsc{GPT-4.1-mini}. The scoring function in \textsc{BM25} measures lexical overlap between chunks using \textit{TF-IDF}-style matching, serving as only a coarse proxy for mutual information. With open-weight \textsc{Qwen-3-Embedding-8B}, we see a similar trend as \textsc{Text-Embedding-3-Large} across all datasets. Here, \textsc{CL--order} consistently outperforms the \textsc{Dense} baseline. Similarly, we observe larger gains for the smaller models, \textsc{GPT-4.1-mini} and \textsc{Qwen-3}. 

\textbf{Traversal Strategy.} We also compare our \textsc{CL--order} ordering strategy against a greedy DFS traversal directly operated on the complete document-chunk graph. The \textsc{DFS} strategy first selects the chunk node most similar to the query, and then proceeds to greedily choose the closest unvisited neighbour from the full document-chunk graph, purely local to chunk $X_t$ at time-step $t$. In contrast, \textsc{CL--order} builds the global dependency structure via a maximum spanning tree, which finds optimal pairwise dependencies across all chunks, before the traversal. Figure \ref{fig:dfsvsclplot} reports this ablation on \textsc{LongQA--MC}. Here, \textsc{CL--order} consistently outperforms \textsc{DFS} across all the three models. \textit{Chow-Liu} captures global dependencies more robustly than local \textsc{DFS}-based chaining. In \textsc{DFS} even a single step toward a highly similar but contextually irrelevant neighbor may lead the traversal away from the optimal path. 

\section{Conclusion}
In this work, we study a dependency-aware chunk ordering strategy for sequential multi-agent reasoning over long contexts. By modeling inter-chunk dependencies using a \textit{Chow--Liu} tree constructed from embedding-based pairwise similarities, our approach (\textsc{CL--order}) derives a breadth-first processing order that keeps related chunks close in the memory update sequence, thereby reducing compression-induced information loss. Empirically, \textsc{CL--order} consistently outperforms default document-chunk ordering and semantic score-based ordering by 10.68\% and 6.89\% relative gains, respectively, in EM-based \textbf{LongQA--MC}. Similarly, \textsc{CL--order} yields relative gains of 5.86\% over the default order and 6.01\% over semantic score-based baseline in Ragas-based \textbf{LongQA, NarrativeQA}, demonstrating that
principled ordering of interdependent document-chunks is a key tool for improving sequential long-context reasoning under memory constraints.

\appendix
\section{Prompt Details}
\label{app:prompt}
This section provides the prompt templates used within the \textit{Chain-of-Agents (CoA)} pipeline. This includes (1) the worker agent prompt, (2) the manager agent prompt, and (3) the task-specific instruction. 

\begin{tcolorbox}[
    colback=RoyalPurple!6,
    colframe=RoyalPurple!48,
    title=\bfseries Worker Prompt,
    fonttitle=\small\sffamily,
    sharp corners,
    boxrule=0.8pt,
    top=1mm, bottom=1mm, left=1mm, right=1mm
]
\begin{lstlisting}[style=prompt]
{chunk}
Here is the summary of the previous source text: {summary_till}
Question: {query}
You need to read current source text and summary of previous source text (if any) and generate a summary to include them both. Later, this summary will be used for other agents to answer the Query, if any. So please write the summary that can include the evidence for answering the Query.
\end{lstlisting}
\end{tcolorbox}

\begin{tcolorbox}[
    colback=RoyalPurple!6,
    colframe=RoyalPurple!48,
    title=\bfseries Manager Prompt,
    fonttitle=\small\sffamily,
    sharp corners,
    boxrule=0.8pt,
    top=1mm, bottom=1mm, left=1mm, right=1mm
]
\begin{lstlisting}[style=prompt]
{task_specific_inst}
The following are given passages. However, the source text is too long and has been summarized. You need to answer based on the summary:
{summary}

Question: {query}
Answer:
\end{lstlisting}
\end{tcolorbox}

\begin{tcolorbox}[
    colback=RoyalPurple!6,
    colframe=RoyalPurple!48,
    title=\bfseries Task-Specific Instruction,
    fonttitle=\small\sffamily,
    sharp corners,
    boxrule=0.8pt,
    top=1mm, bottom=1mm, left=1mm, right=1mm
]
\begin{lstlisting}[style=prompt]
Answer the question based on the context provided. Provide a concise and direct answer to the question. Avoid unnecessary details, explanations, or context. Just the answer is enough.

For example, if the query were "What is the capital of France?", you should answer with "Paris" and not something like "Paris is the capital of France".
\end{lstlisting}
\end{tcolorbox}

\section{Hyperparameter Details}
\label{app:hparam}
\begin{table}[ht]
\centering
\begin{tabular}{lc|lc}
\toprule
\textbf{Hyperparameter} & \textbf{Value} & \textbf{Hyperparameter} & \textbf{Value} \\
\midrule
Per-chunk token limit & 8000 & Qwen embedding size & 4096 \\
Summary token limit   & 8000 & Generation temperature & 0.0 \\
OpenAI embedding size & 3072 & Nucleus sampling (top\_p) & 0.95 \\
\bottomrule
\end{tabular}
\caption{Hyperparameters for the long-context QA problem setting with CoA and Chow--Liu Trees.}
\label{tab:hyperparams}
\end{table}



\begin{thebibliography}{40}
\providecommand{\natexlab}[1]{#1}
\providecommand{\url}[1]{\texttt{#1}}
\expandafter\ifx\csname urlstyle\endcsname\relax
  \providecommand{\doi}[1]{doi: #1}\else
  \providecommand{\doi}{doi: \begingroup \urlstyle{rm}\Url}\fi

\bibitem[Bai et~al.(2024)Bai, Lv, Zhang, Lyu, Tang, Huang, Du, Liu, Zeng, Hou, Dong, Tang, and Li]{bai2023longbench}
Yushi Bai, Xin Lv, Jiajie Zhang, Hongchang Lyu, Jiankai Tang, Zhidian Huang, Zhengxiao Du, Xiao Liu, Aohan Zeng, Lei Hou, Yuxiao Dong, Jie Tang, and Juanzi Li.
\newblock {L}ong{B}ench: A bilingual, multitask benchmark for long context understanding.
\newblock In Lun-Wei Ku, Andre Martins, and Vivek Srikumar (eds.), \emph{Proceedings of the 62nd Annual Meeting of the Association for Computational Linguistics (Volume 1: Long Papers)}, pp.\  3119--3137, Bangkok, Thailand, August 2024. Association for Computational Linguistics.
\newblock \doi{10.18653/v1/2024.acl-long.172}.
\newblock URL \url{https://aclanthology.org/2024.acl-long.172/}.

\bibitem[Beltagy et~al.(2020)Beltagy, Peters, and Cohan]{beltagy2020longformer}
Iz~Beltagy, Matthew~E. Peters, and Arman Cohan.
\newblock Longformer: The long-document transformer, 2020.
\newblock URL \url{https://arxiv.org/abs/2004.05150}.

\bibitem[Brown et~al.(2020)Brown, Mann, Ryder, Subbiah, Kaplan, Dhariwal, Neelakantan, Shyam, Sastry, Askell, Agarwal, Herbert-Voss, Krueger, Henighan, Child, Ramesh, Ziegler, Wu, Winter, Hesse, Chen, Sigler, Litwin, Gray, Chess, Clark, Berner, McCandlish, Radford, Sutskever, and Amodei]{brown2020language}
Tom Brown, Benjamin Mann, Nick Ryder, Melanie Subbiah, Jared~D Kaplan, Prafulla Dhariwal, Arvind Neelakantan, Pranav Shyam, Girish Sastry, Amanda Askell, Sandhini Agarwal, Ariel Herbert-Voss, Gretchen Krueger, Tom Henighan, Rewon Child, Aditya Ramesh, Daniel Ziegler, Jeffrey Wu, Clemens Winter, Chris Hesse, Mark Chen, Eric Sigler, Mateusz Litwin, Scott Gray, Benjamin Chess, Jack Clark, Christopher Berner, Sam McCandlish, Alec Radford, Ilya Sutskever, and Dario Amodei.
\newblock Language models are few-shot learners.
\newblock In H.~Larochelle, M.~Ranzato, R.~Hadsell, M.F. Balcan, and H.~Lin (eds.), \emph{Advances in Neural Information Processing Systems}, volume~33, pp.\  1877--1901. Curran Associates, Inc., 2020.
\newblock URL \url{https://proceedings.neurips.cc/paper_files/paper/2020/file/1457c0d6bfcb4967418bfb8ac142f64a-Paper.pdf}.

\bibitem[Chow \& Liu(1968)Chow and Liu]{chow1968approximating}
C.~Chow and C.~Liu.
\newblock Approximating discrete probability distributions with dependence trees.
\newblock \emph{IEEE Transactions on Information Theory}, 14\penalty0 (3):\penalty0 462--467, 1968.
\newblock \doi{10.1109/TIT.1968.1054142}.

\bibitem[Cui et~al.(2020)Cui, Li, and Zhang]{cui-etal-2020-bert}
Baiyun Cui, Yingming Li, and Zhongfei Zhang.
\newblock {BERT}-enhanced relational sentence ordering network.
\newblock In Bonnie Webber, Trevor Cohn, Yulan He, and Yang Liu (eds.), \emph{Proceedings of the 2020 Conference on Empirical Methods in Natural Language Processing (EMNLP)}, pp.\  6310--6320, Online, November 2020. Association for Computational Linguistics.
\newblock \doi{10.18653/v1/2020.emnlp-main.511}.
\newblock URL \url{https://aclanthology.org/2020.emnlp-main.511/}.

\bibitem[Du et~al.(2024)Du, Li, Torralba, Tenenbaum, and Mordatch]{du2024improving}
Yilun Du, Shuang Li, Antonio Torralba, Joshua~B. Tenenbaum, and Igor Mordatch.
\newblock Improving factuality and reasoning in language models through multiagent debate.
\newblock In \emph{Forty-first International Conference on Machine Learning}, 2024.
\newblock URL \url{https://openreview.net/forum?id=zj7YuTE4t8}.

\bibitem[Es et~al.(2024)Es, James, Espinosa~Anke, and Schockaert]{es2025ragasautomatedevaluationretrieval}
Shahul Es, Jithin James, Luis Espinosa~Anke, and Steven Schockaert.
\newblock {RAGA}s: Automated evaluation of retrieval augmented generation.
\newblock In Nikolaos Aletras and Orphee De~Clercq (eds.), \emph{Proceedings of the 18th Conference of the European Chapter of the Association for Computational Linguistics: System Demonstrations}, pp.\  150--158, St. Julians, Malta, March 2024. Association for Computational Linguistics.
\newblock \doi{10.18653/v1/2024.eacl-demo.16}.
\newblock URL \url{https://aclanthology.org/2024.eacl-demo.16/}.

\bibitem[Grattafiori et~al.(2024)Grattafiori, Dubey, Jauhri, Pandey, Kadian, Al-Dahle, Letman, Mathur, Schelten, Vaughan, Yang, Fan, Goyal, Hartshorn, Yang, Mitra, Sravankumar, Korenev, Hinsvark, Rao, Zhang, Rodriguez, Gregerson, Spataru, Roziere, Biron, Tang, Chern, Caucheteux, Nayak, Bi, Marra, McConnell, Keller, Touret, Wu, Wong, Ferrer, Nikolaidis, Allonsius, Song, Pintz, Livshits, Wyatt, Esiobu, Choudhary, Mahajan, Garcia-Olano, Perino, Hupkes, Lakomkin, AlBadawy, Lobanova, Dinan, Smith, Radenovic, Guzmán, Zhang, Synnaeve, Lee, Anderson, Thattai, Nail, Mialon, Pang, Cucurell, Nguyen, Korevaar, Xu, Touvron, Zarov, Ibarra, Kloumann, Misra, Evtimov, Zhang, Copet, Lee, Geffert, Vranes, Park, Mahadeokar, Shah, van~der Linde, Billock, Hong, Lee, Fu, Chi, Huang, Liu, Wang, Yu, Bitton, Spisak, Park, Rocca, Johnstun, Saxe, Jia, Alwala, Prasad, Upasani, Plawiak, Li, Heafield, Stone, El-Arini, Iyer, Malik, Chiu, Bhalla, Lakhotia, Rantala-Yeary, van~der Maaten, Chen, Tan, Jenkins, Martin, Madaan, Malo, Blecher,
  Landzaat, de~Oliveira, Muzzi, Pasupuleti, Singh, Paluri, Kardas, Tsimpoukelli, Oldham, Rita, Pavlova, Kambadur, Lewis, Si, Singh, Hassan, Goyal, Torabi, Bashlykov, Bogoychev, Chatterji, Zhang, Duchenne, Çelebi, Alrassy, Zhang, Li, Vasic, Weng, Bhargava, Dubal, Krishnan, Koura, Xu, He, Dong, Srinivasan, Ganapathy, Calderer, Cabral, Stojnic, Raileanu, Maheswari, Girdhar, Patel, Sauvestre, Polidoro, Sumbaly, Taylor, Silva, Hou, Wang, Hosseini, Chennabasappa, Singh, Bell, Kim, Edunov, Nie, Narang, Raparthy, Shen, Wan, Bhosale, Zhang, Vandenhende, Batra, Whitman, Sootla, Collot, Gururangan, Borodinsky, Herman, Fowler, Sheasha, Georgiou, Scialom, Speckbacher, Mihaylov, Xiao, Karn, Goswami, Gupta, Ramanathan, Kerkez, Gonguet, Do, Vogeti, Albiero, Petrovic, Chu, Xiong, Fu, Meers, Martinet, Wang, Wang, Tan, Xia, Xie, Jia, Wang, Goldschlag, Gaur, Babaei, Wen, Song, Zhang, Li, Mao, Coudert, Yan, Chen, Papakipos, Singh, Srivastava, Jain, Kelsey, Shajnfeld, Gangidi, Victoria, Goldstand, Menon, Sharma, Boesenberg,
  Baevski, Feinstein, Kallet, Sangani, Teo, Yunus, Lupu, Alvarado, Caples, Gu, Ho, Poulton, Ryan, Ramchandani, Dong, Franco, Goyal, Saraf, Chowdhury, Gabriel, Bharambe, Eisenman, Yazdan, James, Maurer, Leonhardi, Huang, Loyd, Paola, Paranjape, Liu, Wu, Ni, Hancock, Wasti, Spence, Stojkovic, Gamido, Montalvo, Parker, Burton, Mejia, Liu, Wang, Kim, Zhou, Hu, Chu, Cai, Tindal, Feichtenhofer, Gao, Civin, Beaty, Kreymer, Li, Adkins, Xu, Testuggine, David, Parikh, Liskovich, Foss, Wang, Le, Holland, Dowling, Jamil, Montgomery, Presani, Hahn, Wood, Le, Brinkman, Arcaute, Dunbar, Smothers, Sun, Kreuk, Tian, Kokkinos, Ozgenel, Caggioni, Kanayet, Seide, Florez, Schwarz, Badeer, Swee, Halpern, Herman, Sizov, Guangyi, Zhang, Lakshminarayanan, Inan, Shojanazeri, Zou, Wang, Zha, Habeeb, Rudolph, Suk, Aspegren, Goldman, Zhan, Damlaj, Molybog, Tufanov, Leontiadis, Veliche, Gat, Weissman, Geboski, Kohli, Lam, Asher, Gaya, Marcus, Tang, Chan, Zhen, Reizenstein, Teboul, Zhong, Jin, Yang, Cummings, Carvill, Shepard, McPhie,
  Torres, Ginsburg, Wang, Wu, U, Saxena, Khandelwal, Zand, Matosich, Veeraraghavan, Michelena, Li, Jagadeesh, Huang, Chawla, Huang, Chen, Garg, A, Silva, Bell, Zhang, Guo, Yu, Moshkovich, Wehrstedt, Khabsa, Avalani, Bhatt, Mankus, Hasson, Lennie, Reso, Groshev, Naumov, Lathi, Keneally, Liu, Seltzer, Valko, Restrepo, Patel, Vyatskov, Samvelyan, Clark, Macey, Wang, Hermoso, Metanat, Rastegari, Bansal, Santhanam, Parks, White, Bawa, Singhal, Egebo, Usunier, Mehta, Laptev, Dong, Cheng, Chernoguz, Hart, Salpekar, Kalinli, Kent, Parekh, Saab, Balaji, Rittner, Bontrager, Roux, Dollar, Zvyagina, Ratanchandani, Yuvraj, Liang, Alao, Rodriguez, Ayub, Murthy, Nayani, Mitra, Parthasarathy, Li, Hogan, Battey, Wang, Howes, Rinott, Mehta, Siby, Bondu, Datta, Chugh, Hunt, Dhillon, Sidorov, Pan, Mahajan, Verma, Yamamoto, Ramaswamy, Lindsay, Lindsay, Feng, Lin, Zha, Patil, Shankar, Zhang, Zhang, Wang, Agarwal, Sajuyigbe, Chintala, Max, Chen, Kehoe, Satterfield, Govindaprasad, Gupta, Deng, Cho, Virk, Subramanian, Choudhury,
  Goldman, Remez, Glaser, Best, Koehler, Robinson, Li, Zhang, Matthews, Chou, Shaked, Vontimitta, Ajayi, Montanez, Mohan, Kumar, Mangla, Ionescu, Poenaru, Mihailescu, Ivanov, Li, Wang, Jiang, Bouaziz, Constable, Tang, Wu, Wang, Wu, Gao, Kleinman, Chen, Hu, Jia, Qi, Li, Zhang, Zhang, Adi, Nam, Yu, Wang, Zhao, Hao, Qian, Li, He, Rait, DeVito, Rosnbrick, Wen, Yang, Zhao, and Ma]{dubey2024llama}
Aaron Grattafiori, Abhimanyu Dubey, Abhinav Jauhri, Abhinav Pandey, Abhishek Kadian, Ahmad Al-Dahle, Aiesha Letman, Akhil Mathur, Alan Schelten, Alex Vaughan, Amy Yang, Angela Fan, Anirudh Goyal, Anthony Hartshorn, Aobo Yang, Archi Mitra, Archie Sravankumar, Artem Korenev, Arthur Hinsvark, Arun Rao, Aston Zhang, Aurelien Rodriguez, Austen Gregerson, Ava Spataru, Baptiste Roziere, Bethany Biron, Binh Tang, Bobbie Chern, Charlotte Caucheteux, Chaya Nayak, Chloe Bi, Chris Marra, Chris McConnell, Christian Keller, Christophe Touret, Chunyang Wu, Corinne Wong, Cristian~Canton Ferrer, Cyrus Nikolaidis, Damien Allonsius, Daniel Song, Danielle Pintz, Danny Livshits, Danny Wyatt, David Esiobu, Dhruv Choudhary, Dhruv Mahajan, Diego Garcia-Olano, Diego Perino, Dieuwke Hupkes, Egor Lakomkin, Ehab AlBadawy, Elina Lobanova, Emily Dinan, Eric~Michael Smith, Filip Radenovic, Francisco Guzmán, Frank Zhang, Gabriel Synnaeve, Gabrielle Lee, Georgia~Lewis Anderson, Govind Thattai, Graeme Nail, Gregoire Mialon, Guan Pang,
  Guillem Cucurell, Hailey Nguyen, Hannah Korevaar, Hu~Xu, Hugo Touvron, Iliyan Zarov, Imanol~Arrieta Ibarra, Isabel Kloumann, Ishan Misra, Ivan Evtimov, Jack Zhang, Jade Copet, Jaewon Lee, Jan Geffert, Jana Vranes, Jason Park, Jay Mahadeokar, Jeet Shah, Jelmer van~der Linde, Jennifer Billock, Jenny Hong, Jenya Lee, Jeremy Fu, Jianfeng Chi, Jianyu Huang, Jiawen Liu, Jie Wang, Jiecao Yu, Joanna Bitton, Joe Spisak, Jongsoo Park, Joseph Rocca, Joshua Johnstun, Joshua Saxe, Junteng Jia, Kalyan~Vasuden Alwala, Karthik Prasad, Kartikeya Upasani, Kate Plawiak, Ke~Li, Kenneth Heafield, Kevin Stone, Khalid El-Arini, Krithika Iyer, Kshitiz Malik, Kuenley Chiu, Kunal Bhalla, Kushal Lakhotia, Lauren Rantala-Yeary, Laurens van~der Maaten, Lawrence Chen, Liang Tan, Liz Jenkins, Louis Martin, Lovish Madaan, Lubo Malo, Lukas Blecher, Lukas Landzaat, Luke de~Oliveira, Madeline Muzzi, Mahesh Pasupuleti, Mannat Singh, Manohar Paluri, Marcin Kardas, Maria Tsimpoukelli, Mathew Oldham, Mathieu Rita, Maya Pavlova, Melanie Kambadur,
  Mike Lewis, Min Si, Mitesh~Kumar Singh, Mona Hassan, Naman Goyal, Narjes Torabi, Nikolay Bashlykov, Nikolay Bogoychev, Niladri Chatterji, Ning Zhang, Olivier Duchenne, Onur Çelebi, Patrick Alrassy, Pengchuan Zhang, Pengwei Li, Petar Vasic, Peter Weng, Prajjwal Bhargava, Pratik Dubal, Praveen Krishnan, Punit~Singh Koura, Puxin Xu, Qing He, Qingxiao Dong, Ragavan Srinivasan, Raj Ganapathy, Ramon Calderer, Ricardo~Silveira Cabral, Robert Stojnic, Roberta Raileanu, Rohan Maheswari, Rohit Girdhar, Rohit Patel, Romain Sauvestre, Ronnie Polidoro, Roshan Sumbaly, Ross Taylor, Ruan Silva, Rui Hou, Rui Wang, Saghar Hosseini, Sahana Chennabasappa, Sanjay Singh, Sean Bell, Seohyun~Sonia Kim, Sergey Edunov, Shaoliang Nie, Sharan Narang, Sharath Raparthy, Sheng Shen, Shengye Wan, Shruti Bhosale, Shun Zhang, Simon Vandenhende, Soumya Batra, Spencer Whitman, Sten Sootla, Stephane Collot, Suchin Gururangan, Sydney Borodinsky, Tamar Herman, Tara Fowler, Tarek Sheasha, Thomas Georgiou, Thomas Scialom, Tobias Speckbacher,
  Todor Mihaylov, Tong Xiao, Ujjwal Karn, Vedanuj Goswami, Vibhor Gupta, Vignesh Ramanathan, Viktor Kerkez, Vincent Gonguet, Virginie Do, Vish Vogeti, Vítor Albiero, Vladan Petrovic, Weiwei Chu, Wenhan Xiong, Wenyin Fu, Whitney Meers, Xavier Martinet, Xiaodong Wang, Xiaofang Wang, Xiaoqing~Ellen Tan, Xide Xia, Xinfeng Xie, Xuchao Jia, Xuewei Wang, Yaelle Goldschlag, Yashesh Gaur, Yasmine Babaei, Yi~Wen, Yiwen Song, Yuchen Zhang, Yue Li, Yuning Mao, Zacharie~Delpierre Coudert, Zheng Yan, Zhengxing Chen, Zoe Papakipos, Aaditya Singh, Aayushi Srivastava, Abha Jain, Adam Kelsey, Adam Shajnfeld, Adithya Gangidi, Adolfo Victoria, Ahuva Goldstand, Ajay Menon, Ajay Sharma, Alex Boesenberg, Alexei Baevski, Allie Feinstein, Amanda Kallet, Amit Sangani, Amos Teo, Anam Yunus, Andrei Lupu, Andres Alvarado, Andrew Caples, Andrew Gu, Andrew Ho, Andrew Poulton, Andrew Ryan, Ankit Ramchandani, Annie Dong, Annie Franco, Anuj Goyal, Aparajita Saraf, Arkabandhu Chowdhury, Ashley Gabriel, Ashwin Bharambe, Assaf Eisenman, Azadeh
  Yazdan, Beau James, Ben Maurer, Benjamin Leonhardi, Bernie Huang, Beth Loyd, Beto~De Paola, Bhargavi Paranjape, Bing Liu, Bo~Wu, Boyu Ni, Braden Hancock, Bram Wasti, Brandon Spence, Brani Stojkovic, Brian Gamido, Britt Montalvo, Carl Parker, Carly Burton, Catalina Mejia, Ce~Liu, Changhan Wang, Changkyu Kim, Chao Zhou, Chester Hu, Ching-Hsiang Chu, Chris Cai, Chris Tindal, Christoph Feichtenhofer, Cynthia Gao, Damon Civin, Dana Beaty, Daniel Kreymer, Daniel Li, David Adkins, David Xu, Davide Testuggine, Delia David, Devi Parikh, Diana Liskovich, Didem Foss, Dingkang Wang, Duc Le, Dustin Holland, Edward Dowling, Eissa Jamil, Elaine Montgomery, Eleonora Presani, Emily Hahn, Emily Wood, Eric-Tuan Le, Erik Brinkman, Esteban Arcaute, Evan Dunbar, Evan Smothers, Fei Sun, Felix Kreuk, Feng Tian, Filippos Kokkinos, Firat Ozgenel, Francesco Caggioni, Frank Kanayet, Frank Seide, Gabriela~Medina Florez, Gabriella Schwarz, Gada Badeer, Georgia Swee, Gil Halpern, Grant Herman, Grigory Sizov, Guangyi, Zhang, Guna
  Lakshminarayanan, Hakan Inan, Hamid Shojanazeri, Han Zou, Hannah Wang, Hanwen Zha, Haroun Habeeb, Harrison Rudolph, Helen Suk, Henry Aspegren, Hunter Goldman, Hongyuan Zhan, Ibrahim Damlaj, Igor Molybog, Igor Tufanov, Ilias Leontiadis, Irina-Elena Veliche, Itai Gat, Jake Weissman, James Geboski, James Kohli, Janice Lam, Japhet Asher, Jean-Baptiste Gaya, Jeff Marcus, Jeff Tang, Jennifer Chan, Jenny Zhen, Jeremy Reizenstein, Jeremy Teboul, Jessica Zhong, Jian Jin, Jingyi Yang, Joe Cummings, Jon Carvill, Jon Shepard, Jonathan McPhie, Jonathan Torres, Josh Ginsburg, Junjie Wang, Kai Wu, Kam~Hou U, Karan Saxena, Kartikay Khandelwal, Katayoun Zand, Kathy Matosich, Kaushik Veeraraghavan, Kelly Michelena, Keqian Li, Kiran Jagadeesh, Kun Huang, Kunal Chawla, Kyle Huang, Lailin Chen, Lakshya Garg, Lavender A, Leandro Silva, Lee Bell, Lei Zhang, Liangpeng Guo, Licheng Yu, Liron Moshkovich, Luca Wehrstedt, Madian Khabsa, Manav Avalani, Manish Bhatt, Martynas Mankus, Matan Hasson, Matthew Lennie, Matthias Reso, Maxim
  Groshev, Maxim Naumov, Maya Lathi, Meghan Keneally, Miao Liu, Michael~L. Seltzer, Michal Valko, Michelle Restrepo, Mihir Patel, Mik Vyatskov, Mikayel Samvelyan, Mike Clark, Mike Macey, Mike Wang, Miquel~Jubert Hermoso, Mo~Metanat, Mohammad Rastegari, Munish Bansal, Nandhini Santhanam, Natascha Parks, Natasha White, Navyata Bawa, Nayan Singhal, Nick Egebo, Nicolas Usunier, Nikhil Mehta, Nikolay~Pavlovich Laptev, Ning Dong, Norman Cheng, Oleg Chernoguz, Olivia Hart, Omkar Salpekar, Ozlem Kalinli, Parkin Kent, Parth Parekh, Paul Saab, Pavan Balaji, Pedro Rittner, Philip Bontrager, Pierre Roux, Piotr Dollar, Polina Zvyagina, Prashant Ratanchandani, Pritish Yuvraj, Qian Liang, Rachad Alao, Rachel Rodriguez, Rafi Ayub, Raghotham Murthy, Raghu Nayani, Rahul Mitra, Rangaprabhu Parthasarathy, Raymond Li, Rebekkah Hogan, Robin Battey, Rocky Wang, Russ Howes, Ruty Rinott, Sachin Mehta, Sachin Siby, Sai~Jayesh Bondu, Samyak Datta, Sara Chugh, Sara Hunt, Sargun Dhillon, Sasha Sidorov, Satadru Pan, Saurabh Mahajan,
  Saurabh Verma, Seiji Yamamoto, Sharadh Ramaswamy, Shaun Lindsay, Shaun Lindsay, Sheng Feng, Shenghao Lin, Shengxin~Cindy Zha, Shishir Patil, Shiva Shankar, Shuqiang Zhang, Shuqiang Zhang, Sinong Wang, Sneha Agarwal, Soji Sajuyigbe, Soumith Chintala, Stephanie Max, Stephen Chen, Steve Kehoe, Steve Satterfield, Sudarshan Govindaprasad, Sumit Gupta, Summer Deng, Sungmin Cho, Sunny Virk, Suraj Subramanian, Sy~Choudhury, Sydney Goldman, Tal Remez, Tamar Glaser, Tamara Best, Thilo Koehler, Thomas Robinson, Tianhe Li, Tianjun Zhang, Tim Matthews, Timothy Chou, Tzook Shaked, Varun Vontimitta, Victoria Ajayi, Victoria Montanez, Vijai Mohan, Vinay~Satish Kumar, Vishal Mangla, Vlad Ionescu, Vlad Poenaru, Vlad~Tiberiu Mihailescu, Vladimir Ivanov, Wei Li, Wenchen Wang, Wenwen Jiang, Wes Bouaziz, Will Constable, Xiaocheng Tang, Xiaojian Wu, Xiaolan Wang, Xilun Wu, Xinbo Gao, Yaniv Kleinman, Yanjun Chen, Ye~Hu, Ye~Jia, Ye~Qi, Yenda Li, Yilin Zhang, Ying Zhang, Yossi Adi, Youngjin Nam, Yu, Wang, Yu~Zhao, Yuchen Hao, Yundi
  Qian, Yunlu Li, Yuzi He, Zach Rait, Zachary DeVito, Zef Rosnbrick, Zhaoduo Wen, Zhenyu Yang, Zhiwei Zhao, and Zhiyu Ma.
\newblock The llama 3 herd of models, 2024.
\newblock URL \url{https://arxiv.org/abs/2407.21783}.

\bibitem[Gupta et~al.(2025)Gupta, Gowaikar, Iyer, Shiragur, Bairi, Maurya, Maiti, Damle, and Gupta]{gupta2025cosmir}
Naman Gupta, Shreeyash Gowaikar, Arun Iyer, Kirankumar Shiragur, Ramakrishna~B Bairi, Rishikesh Maurya, Ritabrata Maiti, Sankarshan Damle, and Shachee~Mishra Gupta.
\newblock {COSMIR}: Chain orchestrated structured memory for iterative reasoning over long context.
\newblock In \emph{First Workshop on Foundations of Reasoning in Language Models}, 2025.
\newblock URL \url{https://openreview.net/forum?id=rQ1hAQqrd4}.

\bibitem[Guu et~al.(2020)Guu, Lee, Tung, Pasupat, and Chang]{pmlr-v119-guu20a}
Kelvin Guu, Kenton Lee, Zora Tung, Panupong Pasupat, and Mingwei Chang.
\newblock Retrieval augmented language model pre-training.
\newblock In Hal~Daumé III and Aarti Singh (eds.), \emph{Proceedings of the 37th International Conference on Machine Learning}, volume 119 of \emph{Proceedings of Machine Learning Research}, pp.\  3929--3938. PMLR, 13--18 Jul 2020.
\newblock URL \url{https://proceedings.mlr.press/v119/guu20a.html}.

\bibitem[Hong et~al.(2024)Hong, Zhuge, Chen, Zheng, Cheng, Wang, Zhang, Wang, Yau, Lin, Zhou, Ran, Xiao, Wu, and Schmidhuber]{hong2023metagpt}
Sirui Hong, Mingchen Zhuge, Jonathan Chen, Xiawu Zheng, Yuheng Cheng, Jinlin Wang, Ceyao Zhang, Zili Wang, Steven Ka~Shing Yau, Zijuan Lin, Liyang Zhou, Chenyu Ran, Lingfeng Xiao, Chenglin Wu, and J{\"u}rgen Schmidhuber.
\newblock Meta{GPT}: Meta programming for a multi-agent collaborative framework.
\newblock In \emph{The Twelfth International Conference on Learning Representations}, 2024.
\newblock URL \url{https://openreview.net/forum?id=VtmBAGCN7o}.

\bibitem[Hsieh et~al.(2024)Hsieh, Sun, Kriman, Acharya, Rekesh, Jia, and Ginsburg]{hsieh2024ruler}
Cheng-Ping Hsieh, Simeng Sun, Samuel Kriman, Shantanu Acharya, Dima Rekesh, Fei Jia, and Boris Ginsburg.
\newblock {RULER}: What{\textquoteright}s the real context size of your long-context language models?
\newblock In \emph{First Conference on Language Modeling}, 2024.
\newblock URL \url{https://openreview.net/forum?id=kIoBbc76Sy}.

\bibitem[Izacard \& Grave(2021)Izacard and Grave]{izacard2021leveraging}
Gautier Izacard and Edouard Grave.
\newblock Leveraging passage retrieval with generative models for open domain question answering.
\newblock In Paola Merlo, Jorg Tiedemann, and Reut Tsarfaty (eds.), \emph{Proceedings of the 16th Conference of the European Chapter of the Association for Computational Linguistics: Main Volume}, pp.\  874--880, Online, April 2021. Association for Computational Linguistics.
\newblock \doi{10.18653/v1/2021.eacl-main.74}.
\newblock URL \url{https://aclanthology.org/2021.eacl-main.74/}.

\bibitem[Jiao et~al.(2016)Jiao, Han, and Weissman]{han2023beyond}
Jiantao Jiao, Yanjun Han, and Tsachy Weissman.
\newblock Beyond maximum likelihood: Boosting the chow-liu algorithm for large alphabets.
\newblock In \emph{2016 50th Asilomar Conference on Signals, Systems and Computers}, pp.\  321--325, 2016.
\newblock \doi{10.1109/ACSSC.2016.7869051}.

\bibitem[Khattab et~al.(2021)Khattab, Potts, and Zaharia]{khattab2021baleen}
Omar Khattab, Christopher Potts, and Matei Zaharia.
\newblock Baleen: Robust multi-hop reasoning at scale via condensed retrieval.
\newblock In A.~Beygelzimer, Y.~Dauphin, P.~Liang, and J.~Wortman Vaughan (eds.), \emph{Advances in Neural Information Processing Systems}, 2021.
\newblock URL \url{https://openreview.net/forum?id=Ghk0AJ8XtVx}.

\bibitem[Ko{\v{c}}isk{\'y} et~al.(2018)Ko{\v{c}}isk{\'y}, Schwarz, Blunsom, Dyer, Hermann, Melis, and Grefenstette]{kocisky-etal-2018-narrativeqa}
Tom{\'a}{\v{s}} Ko{\v{c}}isk{\'y}, Jonathan Schwarz, Phil Blunsom, Chris Dyer, Karl~Moritz Hermann, G{\'a}bor Melis, and Edward Grefenstette.
\newblock The {N}arrative{QA} reading comprehension challenge.
\newblock \emph{Transactions of the Association for Computational Linguistics}, 6:\penalty0 317--328, 2018.
\newblock \doi{10.1162/tacl_a_00023}.
\newblock URL \url{https://aclanthology.org/Q18-1023/}.

\bibitem[Lewis et~al.(2020)Lewis, Perez, Piktus, Petroni, Karpukhin, Goyal, K\"{u}ttler, Lewis, Yih, Rockt\"{a}schel, Riedel, and Kiela]{NEURIPS2020_6b493230}
Patrick Lewis, Ethan Perez, Aleksandra Piktus, Fabio Petroni, Vladimir Karpukhin, Naman Goyal, Heinrich K\"{u}ttler, Mike Lewis, Wen-tau Yih, Tim Rockt\"{a}schel, Sebastian Riedel, and Douwe Kiela.
\newblock Retrieval-augmented generation for knowledge-intensive nlp tasks.
\newblock In H.~Larochelle, M.~Ranzato, R.~Hadsell, M.F. Balcan, and H.~Lin (eds.), \emph{Advances in Neural Information Processing Systems}, volume~33, pp.\  9459--9474. Curran Associates, Inc., 2020.
\newblock URL \url{https://proceedings.neurips.cc/paper_files/paper/2020/file/6b493230205f780e1bc26945df7481e5-Paper.pdf}.

\bibitem[Lin(2004)]{lin-2004-rouge}
Chin-Yew Lin.
\newblock {ROUGE}: A package for automatic evaluation of summaries.
\newblock In \emph{Text Summarization Branches Out}, pp.\  74--81, Barcelona, Spain, July 2004. Association for Computational Linguistics.
\newblock URL \url{https://aclanthology.org/W04-1013/}.

\bibitem[Liu et~al.(2024)Liu, Lin, Hewitt, Paranjape, Bevilacqua, Petroni, and Liang]{liu2024lost}
Nelson~F. Liu, Kevin Lin, John Hewitt, Ashwin Paranjape, Michele Bevilacqua, Fabio Petroni, and Percy Liang.
\newblock Lost in the middle: How language models use long contexts.
\newblock \emph{Transactions of the Association for Computational Linguistics}, 12:\penalty0 157--173, 2024.
\newblock \doi{10.1162/tacl_a_00638}.
\newblock URL \url{https://aclanthology.org/2024.tacl-1.9/}.

\bibitem[Meila \& Jordan(2001)Meila and Jordan]{10.1162/153244301753344605}
Marina Meila and Michael~I. Jordan.
\newblock Learning with mixtures of trees.
\newblock \emph{J. Mach. Learn. Res.}, 1:\penalty0 1–48, September 2001.
\newblock ISSN 1532-4435.
\newblock \doi{10.1162/153244301753344605}.
\newblock URL \url{https://doi.org/10.1162/153244301753344605}.

\bibitem[OpenAI(2024{\natexlab{a}})]{openai2024gpt41}
OpenAI.
\newblock Gpt-4.1 technical overview.
\newblock \url{https://platform.openai.com/docs/models}, 2024{\natexlab{a}}.

\bibitem[OpenAI(2024{\natexlab{b}})]{openai2024textembedding3large}
OpenAI.
\newblock text-embedding-3-large model documentation.
\newblock \url{https://platform.openai.com/docs/models/text-embedding-3-large}, 2024{\natexlab{b}}.

\bibitem[OpenAI et~al.(2024)OpenAI, Achiam, Adler, Agarwal, Ahmad, Akkaya, Aleman, Almeida, Altenschmidt, Altman, Anadkat, Avila, Babuschkin, Balaji, Balcom, Baltescu, Bao, Bavarian, Belgum, Bello, Berdine, Bernadett-Shapiro, Berner, Bogdonoff, Boiko, Boyd, Brakman, Brockman, Brooks, Brundage, Button, Cai, Campbell, Cann, Carey, Carlson, Carmichael, Chan, Chang, Chantzis, Chen, Chen, Chen, Chen, Chen, Chess, Cho, Chu, Chung, Cummings, Currier, Dai, Decareaux, Degry, Deutsch, Deville, Dhar, Dohan, Dowling, Dunning, Ecoffet, Eleti, Eloundou, Farhi, Fedus, Felix, Fishman, Forte, Fulford, Gao, Georges, Gibson, Goel, Gogineni, Goh, Gontijo-Lopes, Gordon, Grafstein, Gray, Greene, Gross, Gu, Guo, Hallacy, Han, Harris, He, Heaton, Heidecke, Hesse, Hickey, Hickey, Hoeschele, Houghton, Hsu, Hu, Hu, Huizinga, Jain, Jain, Jang, Jiang, Jiang, Jin, Jin, Jomoto, Jonn, Jun, Kaftan, Łukasz Kaiser, Kamali, Kanitscheider, Keskar, Khan, Kilpatrick, Kim, Kim, Kim, Kirchner, Kiros, Knight, Kokotajlo, Łukasz Kondraciuk, Kondrich,
  Konstantinidis, Kosic, Krueger, Kuo, Lampe, Lan, Lee, Leike, Leung, Levy, Li, Lim, Lin, Lin, Litwin, Lopez, Lowe, Lue, Makanju, Malfacini, Manning, Markov, Markovski, Martin, Mayer, Mayne, McGrew, McKinney, McLeavey, McMillan, McNeil, Medina, Mehta, Menick, Metz, Mishchenko, Mishkin, Monaco, Morikawa, Mossing, Mu, Murati, Murk, Mély, Nair, Nakano, Nayak, Neelakantan, Ngo, Noh, Ouyang, O'Keefe, Pachocki, Paino, Palermo, Pantuliano, Parascandolo, Parish, Parparita, Passos, Pavlov, Peng, Perelman, de~Avila Belbute~Peres, Petrov, de~Oliveira~Pinto, Michael, Pokorny, Pokrass, Pong, Powell, Power, Power, Proehl, Puri, Radford, Rae, Ramesh, Raymond, Real, Rimbach, Ross, Rotsted, Roussez, Ryder, Saltarelli, Sanders, Santurkar, Sastry, Schmidt, Schnurr, Schulman, Selsam, Sheppard, Sherbakov, Shieh, Shoker, Shyam, Sidor, Sigler, Simens, Sitkin, Slama, Sohl, Sokolowsky, Song, Staudacher, Such, Summers, Sutskever, Tang, Tezak, Thompson, Tillet, Tootoonchian, Tseng, Tuggle, Turley, Tworek, Uribe, Vallone, Vijayvergiya,
  Voss, Wainwright, Wang, Wang, Wang, Ward, Wei, Weinmann, Welihinda, Welinder, Weng, Weng, Wiethoff, Willner, Winter, Wolrich, Wong, Workman, Wu, Wu, Wu, Xiao, Xu, Yoo, Yu, Yuan, Zaremba, Zellers, Zhang, Zhang, Zhao, Zheng, Zhuang, Zhuk, and Zoph]{openai2024gpt4technicalreport}
OpenAI, Josh Achiam, Steven Adler, Sandhini Agarwal, Lama Ahmad, Ilge Akkaya, Florencia~Leoni Aleman, Diogo Almeida, Janko Altenschmidt, Sam Altman, Shyamal Anadkat, Red Avila, Igor Babuschkin, Suchir Balaji, Valerie Balcom, Paul Baltescu, Haiming Bao, Mohammad Bavarian, Jeff Belgum, Irwan Bello, Jake Berdine, Gabriel Bernadett-Shapiro, Christopher Berner, Lenny Bogdonoff, Oleg Boiko, Madelaine Boyd, Anna-Luisa Brakman, Greg Brockman, Tim Brooks, Miles Brundage, Kevin Button, Trevor Cai, Rosie Campbell, Andrew Cann, Brittany Carey, Chelsea Carlson, Rory Carmichael, Brooke Chan, Che Chang, Fotis Chantzis, Derek Chen, Sully Chen, Ruby Chen, Jason Chen, Mark Chen, Ben Chess, Chester Cho, Casey Chu, Hyung~Won Chung, Dave Cummings, Jeremiah Currier, Yunxing Dai, Cory Decareaux, Thomas Degry, Noah Deutsch, Damien Deville, Arka Dhar, David Dohan, Steve Dowling, Sheila Dunning, Adrien Ecoffet, Atty Eleti, Tyna Eloundou, David Farhi, Liam Fedus, Niko Felix, Simón~Posada Fishman, Juston Forte, Isabella Fulford, Leo
  Gao, Elie Georges, Christian Gibson, Vik Goel, Tarun Gogineni, Gabriel Goh, Rapha Gontijo-Lopes, Jonathan Gordon, Morgan Grafstein, Scott Gray, Ryan Greene, Joshua Gross, Shixiang~Shane Gu, Yufei Guo, Chris Hallacy, Jesse Han, Jeff Harris, Yuchen He, Mike Heaton, Johannes Heidecke, Chris Hesse, Alan Hickey, Wade Hickey, Peter Hoeschele, Brandon Houghton, Kenny Hsu, Shengli Hu, Xin Hu, Joost Huizinga, Shantanu Jain, Shawn Jain, Joanne Jang, Angela Jiang, Roger Jiang, Haozhun Jin, Denny Jin, Shino Jomoto, Billie Jonn, Heewoo Jun, Tomer Kaftan, Łukasz Kaiser, Ali Kamali, Ingmar Kanitscheider, Nitish~Shirish Keskar, Tabarak Khan, Logan Kilpatrick, Jong~Wook Kim, Christina Kim, Yongjik Kim, Jan~Hendrik Kirchner, Jamie Kiros, Matt Knight, Daniel Kokotajlo, Łukasz Kondraciuk, Andrew Kondrich, Aris Konstantinidis, Kyle Kosic, Gretchen Krueger, Vishal Kuo, Michael Lampe, Ikai Lan, Teddy Lee, Jan Leike, Jade Leung, Daniel Levy, Chak~Ming Li, Rachel Lim, Molly Lin, Stephanie Lin, Mateusz Litwin, Theresa Lopez, Ryan
  Lowe, Patricia Lue, Anna Makanju, Kim Malfacini, Sam Manning, Todor Markov, Yaniv Markovski, Bianca Martin, Katie Mayer, Andrew Mayne, Bob McGrew, Scott~Mayer McKinney, Christine McLeavey, Paul McMillan, Jake McNeil, David Medina, Aalok Mehta, Jacob Menick, Luke Metz, Andrey Mishchenko, Pamela Mishkin, Vinnie Monaco, Evan Morikawa, Daniel Mossing, Tong Mu, Mira Murati, Oleg Murk, David Mély, Ashvin Nair, Reiichiro Nakano, Rajeev Nayak, Arvind Neelakantan, Richard Ngo, Hyeonwoo Noh, Long Ouyang, Cullen O'Keefe, Jakub Pachocki, Alex Paino, Joe Palermo, Ashley Pantuliano, Giambattista Parascandolo, Joel Parish, Emy Parparita, Alex Passos, Mikhail Pavlov, Andrew Peng, Adam Perelman, Filipe de~Avila Belbute~Peres, Michael Petrov, Henrique~Ponde de~Oliveira~Pinto, Michael, Pokorny, Michelle Pokrass, Vitchyr~H. Pong, Tolly Powell, Alethea Power, Boris Power, Elizabeth Proehl, Raul Puri, Alec Radford, Jack Rae, Aditya Ramesh, Cameron Raymond, Francis Real, Kendra Rimbach, Carl Ross, Bob Rotsted, Henri Roussez,
  Nick Ryder, Mario Saltarelli, Ted Sanders, Shibani Santurkar, Girish Sastry, Heather Schmidt, David Schnurr, John Schulman, Daniel Selsam, Kyla Sheppard, Toki Sherbakov, Jessica Shieh, Sarah Shoker, Pranav Shyam, Szymon Sidor, Eric Sigler, Maddie Simens, Jordan Sitkin, Katarina Slama, Ian Sohl, Benjamin Sokolowsky, Yang Song, Natalie Staudacher, Felipe~Petroski Such, Natalie Summers, Ilya Sutskever, Jie Tang, Nikolas Tezak, Madeleine~B. Thompson, Phil Tillet, Amin Tootoonchian, Elizabeth Tseng, Preston Tuggle, Nick Turley, Jerry Tworek, Juan Felipe~Cerón Uribe, Andrea Vallone, Arun Vijayvergiya, Chelsea Voss, Carroll Wainwright, Justin~Jay Wang, Alvin Wang, Ben Wang, Jonathan Ward, Jason Wei, CJ~Weinmann, Akila Welihinda, Peter Welinder, Jiayi Weng, Lilian Weng, Matt Wiethoff, Dave Willner, Clemens Winter, Samuel Wolrich, Hannah Wong, Lauren Workman, Sherwin Wu, Jeff Wu, Michael Wu, Kai Xiao, Tao Xu, Sarah Yoo, Kevin Yu, Qiming Yuan, Wojciech Zaremba, Rowan Zellers, Chong Zhang, Marvin Zhang, Shengjia
  Zhao, Tianhao Zheng, Juntang Zhuang, William Zhuk, and Barret Zoph.
\newblock Gpt-4 technical report, 2024.
\newblock URL \url{https://arxiv.org/abs/2303.08774}.

\bibitem[Pour et~al.(2020)Pour, Razavi, and Faili]{golestani2021sentenceordering}
Melika~Golestani Pour, Seyedeh~Zahra Razavi, and Heshaam Faili.
\newblock A new sentence ordering method using bert pretrained model.
\newblock In \emph{2020 11th International Conference on Information and Knowledge Technology (IKT)}, pp.\  132--138, 2020.
\newblock \doi{10.1109/IKT51791.2020.9345618}.

\bibitem[Prabhumoye et~al.(2020)Prabhumoye, Salakhutdinov, and Black]{prabhumoye-etal-2020-topological}
Shrimai Prabhumoye, Ruslan Salakhutdinov, and Alan~W Black.
\newblock Topological sort for sentence ordering.
\newblock In Dan Jurafsky, Joyce Chai, Natalie Schluter, and Joel Tetreault (eds.), \emph{Proceedings of the 58th Annual Meeting of the Association for Computational Linguistics}, pp.\  2783--2792, Online, July 2020. Association for Computational Linguistics.
\newblock \doi{10.18653/v1/2020.acl-main.248}.
\newblock URL \url{https://aclanthology.org/2020.acl-main.248/}.

\bibitem[Press et~al.(2022)Press, Smith, and Lewis]{press2022train}
Ofir Press, Noah~A. Smith, and Mike Lewis.
\newblock Train short, test long: Attention with linear biases enables input length extrapolation, 2022.
\newblock URL \url{https://arxiv.org/abs/2108.12409}.

\bibitem[Rahman et~al.(2014)Rahman, Kothalkar, and Gogate]{rahman2014cutset}
Tahrima Rahman, Prasanna Kothalkar, and Vibhav Gogate.
\newblock Cutset networks: A simple, tractable, and scalable approach for improving the accuracy of chow-liu trees.
\newblock In Toon Calders, Floriana Esposito, Eyke H{\"u}llermeier, and Rosa Meo (eds.), \emph{Machine Learning and Knowledge Discovery in Databases}, pp.\  630--645, Berlin, Heidelberg, 2014. Springer Berlin Heidelberg.
\newblock ISBN 978-3-662-44851-9.

\bibitem[Ram et~al.(2023)Ram, Levine, Dalmedigos, Muhlgay, Shashua, Leyton-Brown, and Shoham]{ram-etal-2023-context}
Ori Ram, Yoav Levine, Itay Dalmedigos, Dor Muhlgay, Amnon Shashua, Kevin Leyton-Brown, and Yoav Shoham.
\newblock In-context retrieval-augmented language models.
\newblock \emph{Transactions of the Association for Computational Linguistics}, 11:\penalty0 1316--1331, 2023.
\newblock \doi{10.1162/tacl_a_00605}.
\newblock URL \url{https://aclanthology.org/2023.tacl-1.75/}.

\bibitem[Song et~al.(2024)Song, Oh, Mo, Kim, Yun, Ha, and Shin]{song2024hierarchical}
Woomin Song, Seunghyuk Oh, Sangwoo Mo, Jaehyung Kim, Sukmin Yun, Jung-Woo Ha, and Jinwoo Shin.
\newblock Hierarchical context merging: Better long context understanding for pre-trained {LLM}s.
\newblock In \emph{The Twelfth International Conference on Learning Representations}, 2024.
\newblock URL \url{https://openreview.net/forum?id=ulaUJFd96G}.

\bibitem[Su et~al.(2024)Su, Ahmed, Lu, Pan, Bo, and Liu]{su2024roformer}
Jianlin Su, Murtadha Ahmed, Yu~Lu, Shengfeng Pan, Wen Bo, and Yunfeng Liu.
\newblock Roformer: Enhanced transformer with rotary position embedding.
\newblock \emph{Neurocomput.}, 568\penalty0 (C), February 2024.
\newblock ISSN 0925-2312.
\newblock \doi{10.1016/j.neucom.2023.127063}.
\newblock URL \url{https://doi.org/10.1016/j.neucom.2023.127063}.

\bibitem[Team et~al.(2024)Team, Georgiev, Lei, Burnell, Bai, Gulati, Tanzer, Vincent, Pan, Wang, Mariooryad, Ding, Geng, Alcober, Frostig, Omernick, Walker, Paduraru, Sorokin, Tacchetti, Gaffney, Daruki, Sercinoglu, Gleicher, Love, Voigtlaender, Jain, Surita, Mohamed, Blevins, Ahn, Zhu, Kawintiranon, Firat, Gu, Zhang, Rahtz, Faruqui, Clay, Gilmer, Co-Reyes, Penchev, Zhu, Morioka, Hui, Haridasan, Campos, Mahdieh, Guo, Hassan, Kilgour, Vezer, Cheng, de~Liedekerke, Goyal, Barham, Strouse, Noury, Adler, Sundararajan, Vikram, Lepikhin, Paganini, Garcia, Yang, Valter, Trebacz, Vodrahalli, Asawaroengchai, Ring, Kalb, Soares, Brahma, Steiner, Yu, Mentzer, He, Gonzalez, Xu, Kaufman, Shafey, Oh, Hennigan, van~den Driessche, Odoom, Lucic, Roelofs, Lall, Marathe, Chan, Ontanon, He, Teplyashin, Lai, Crone, Damoc, Ho, Riedel, Lenc, Yeh, Chowdhery, Xu, Kazemi, Amid, Petrushkina, Swersky, Khodaei, Chen, Larkin, Pinto, Yan, Badia, Patil, Hansen, Orr, Arnold, Grimstad, Dai, Douglas, Sinha, Yadav, Chen, Gribovskaya, Austin,
  Zhao, Patel, Komarek, Austin, Borgeaud, Friso, Goyal, Caine, Cao, Chung, Lamm, Barth-Maron, Kagohara, Olszewska, Chen, Shivakumar, Agarwal, Godhia, Rajwar, Snaider, Dotiwalla, Liu, Barua, Ungureanu, Zhang, Batsaikhan, Wirth, Qin, Danihelka, Doshi, Chadwick, Chen, Jain, Le, Kar, Gurumurthy, Li, Sang, Liu, Lamprou, Munoz, Lintz, Mehta, Howard, Reynolds, Aroyo, Wang, Blanco, Cassirer, Griffith, Das, Lee, Sygnowski, Fisher, Besley, Powell, Ahmed, Paulus, Reitter, Borsos, Joshi, Pope, Hand, Selo, Jain, Sethi, Goel, Makino, May, Yang, Schalkwyk, Butterfield, Hauth, Goldin, Hawkins, Senter, Brin, Woodman, Ritter, Noland, Giang, Bolina, Lee, Blyth, Mackinnon, Reid, Sarvana, Silver, et~al.]{geminiteam2024gemini15unlockingmultimodal}
Gemini Team, Petko Georgiev, Ving~Ian Lei, Ryan Burnell, Libin Bai, Anmol Gulati, Garrett Tanzer, Damien Vincent, Zhufeng Pan, Shibo Wang, Soroosh Mariooryad, Yifan Ding, Xinyang Geng, Fred Alcober, Roy Frostig, Mark Omernick, Lexi Walker, Cosmin Paduraru, Christina Sorokin, Andrea Tacchetti, Colin Gaffney, Samira Daruki, Olcan Sercinoglu, Zach Gleicher, Juliette Love, Paul Voigtlaender, Rohan Jain, Gabriela Surita, Kareem Mohamed, Rory Blevins, Junwhan Ahn, Tao Zhu, Kornraphop Kawintiranon, Orhan Firat, Yiming Gu, Yujing Zhang, Matthew Rahtz, Manaal Faruqui, Natalie Clay, Justin Gilmer, JD~Co-Reyes, Ivo Penchev, Rui Zhu, Nobuyuki Morioka, Kevin Hui, Krishna Haridasan, Victor Campos, Mahdis Mahdieh, Mandy Guo, Samer Hassan, Kevin Kilgour, Arpi Vezer, Heng-Tze Cheng, Raoul de~Liedekerke, Siddharth Goyal, Paul Barham, DJ~Strouse, Seb Noury, Jonas Adler, Mukund Sundararajan, Sharad Vikram, Dmitry Lepikhin, Michela Paganini, Xavier Garcia, Fan Yang, Dasha Valter, Maja Trebacz, Kiran Vodrahalli, Chulayuth
  Asawaroengchai, Roman Ring, Norbert Kalb, Livio~Baldini Soares, Siddhartha Brahma, David Steiner, Tianhe Yu, Fabian Mentzer, Antoine He, Lucas Gonzalez, Bibo Xu, Raphael~Lopez Kaufman, Laurent~El Shafey, Junhyuk Oh, Tom Hennigan, George van~den Driessche, Seth Odoom, Mario Lucic, Becca Roelofs, Sid Lall, Amit Marathe, Betty Chan, Santiago Ontanon, Luheng He, Denis Teplyashin, Jonathan Lai, Phil Crone, Bogdan Damoc, Lewis Ho, Sebastian Riedel, Karel Lenc, Chih-Kuan Yeh, Aakanksha Chowdhery, Yang Xu, Mehran Kazemi, Ehsan Amid, Anastasia Petrushkina, Kevin Swersky, Ali Khodaei, Gowoon Chen, Chris Larkin, Mario Pinto, Geng Yan, Adria~Puigdomenech Badia, Piyush Patil, Steven Hansen, Dave Orr, Sebastien M.~R. Arnold, Jordan Grimstad, Andrew Dai, Sholto Douglas, Rishika Sinha, Vikas Yadav, Xi~Chen, Elena Gribovskaya, Jacob Austin, Jeffrey Zhao, Kaushal Patel, Paul Komarek, Sophia Austin, Sebastian Borgeaud, Linda Friso, Abhimanyu Goyal, Ben Caine, Kris Cao, Da-Woon Chung, Matthew Lamm, Gabe Barth-Maron, Thais
  Kagohara, Kate Olszewska, Mia Chen, Kaushik Shivakumar, Rishabh Agarwal, Harshal Godhia, Ravi Rajwar, Javier Snaider, Xerxes Dotiwalla, Yuan Liu, Aditya Barua, Victor Ungureanu, Yuan Zhang, Bat-Orgil Batsaikhan, Mateo Wirth, James Qin, Ivo Danihelka, Tulsee Doshi, Martin Chadwick, Jilin Chen, Sanil Jain, Quoc Le, Arjun Kar, Madhu Gurumurthy, Cheng Li, Ruoxin Sang, Fangyu Liu, Lampros Lamprou, Rich Munoz, Nathan Lintz, Harsh Mehta, Heidi Howard, Malcolm Reynolds, Lora Aroyo, Quan Wang, Lorenzo Blanco, Albin Cassirer, Jordan Griffith, Dipanjan Das, Stephan Lee, Jakub Sygnowski, Zach Fisher, James Besley, Richard Powell, Zafarali Ahmed, Dominik Paulus, David Reitter, Zalan Borsos, Rishabh Joshi, Aedan Pope, Steven Hand, Vittorio Selo, Vihan Jain, Nikhil Sethi, Megha Goel, Takaki Makino, Rhys May, Zhen Yang, Johan Schalkwyk, Christina Butterfield, Anja Hauth, Alex Goldin, Will Hawkins, Evan Senter, Sergey Brin, Oliver Woodman, Marvin Ritter, Eric Noland, Minh Giang, Vijay Bolina, Lisa Lee, Tim Blyth, Ian
  Mackinnon, Machel Reid, Obaid Sarvana, David Silver, et~al.
\newblock Gemini 1.5: Unlocking multimodal understanding across millions of tokens of context, 2024.
\newblock URL \url{https://arxiv.org/abs/2403.05530}.

\bibitem[Touvron et~al.(2023)Touvron, Lavril, Izacard, Martinet, Lachaux, Lacroix, Rozière, Goyal, Hambro, Azhar, Rodriguez, Joulin, Grave, and Lample]{touvron2023llama}
Hugo Touvron, Thibaut Lavril, Gautier Izacard, Xavier Martinet, Marie-Anne Lachaux, Timothée Lacroix, Baptiste Rozière, Naman Goyal, Eric Hambro, Faisal Azhar, Aurelien Rodriguez, Armand Joulin, Edouard Grave, and Guillaume Lample.
\newblock Llama: Open and efficient foundation language models, 2023.
\newblock URL \url{https://arxiv.org/abs/2302.13971}.

\bibitem[Wei et~al.(2022)Wei, Wang, Schuurmans, Bosma, brian ichter, Xia, Chi, Le, and Zhou]{wei2022chain}
Jason Wei, Xuezhi Wang, Dale Schuurmans, Maarten Bosma, brian ichter, Fei Xia, Ed~H. Chi, Quoc~V Le, and Denny Zhou.
\newblock Chain of thought prompting elicits reasoning in large language models.
\newblock In Alice~H. Oh, Alekh Agarwal, Danielle Belgrave, and Kyunghyun Cho (eds.), \emph{Advances in Neural Information Processing Systems}, 2022.
\newblock URL \url{https://openreview.net/forum?id=_VjQlMeSB_J}.

\bibitem[Xu et~al.(2026)Xu, Zhu, WANG, Wang, Athiwaratkun, Wang, Zou, and Zhang]{xu2026when}
Zach Xu, Shang Zhu, Jue WANG, Junlin Wang, Ben Athiwaratkun, Chi Wang, James Zou, and Ce~Zhang.
\newblock When does divide and conquer work for long context {LLM}? a noise decomposition framework.
\newblock In \emph{The Fourteenth International Conference on Learning Representations}, 2026.
\newblock URL \url{https://openreview.net/forum?id=ddQFUuHDDt}.

\bibitem[Yang et~al.(2025)Yang, Li, Yang, Zhang, Hui, Zheng, Yu, Gao, Huang, Lv, Zheng, Liu, Zhou, Huang, Hu, Ge, Wei, Lin, Tang, Yang, Tu, Zhang, Yang, Yang, Zhou, Zhou, Lin, Dang, Bao, Yang, Yu, Deng, Li, Xue, Li, Zhang, Wang, Zhu, Men, Gao, Liu, Luo, Li, Tang, Yin, Ren, Wang, Zhang, Ren, Fan, Su, Zhang, Zhang, Wan, Liu, Wang, Cui, Zhang, Zhou, and Qiu]{yang2025qwen3technicalreport}
An~Yang, Anfeng Li, Baosong Yang, Beichen Zhang, Binyuan Hui, Bo~Zheng, Bowen Yu, Chang Gao, Chengen Huang, Chenxu Lv, Chujie Zheng, Dayiheng Liu, Fan Zhou, Fei Huang, Feng Hu, Hao Ge, Haoran Wei, Huan Lin, Jialong Tang, Jian Yang, Jianhong Tu, Jianwei Zhang, Jianxin Yang, Jiaxi Yang, Jing Zhou, Jingren Zhou, Junyang Lin, Kai Dang, Keqin Bao, Kexin Yang, Le~Yu, Lianghao Deng, Mei Li, Mingfeng Xue, Mingze Li, Pei Zhang, Peng Wang, Qin Zhu, Rui Men, Ruize Gao, Shixuan Liu, Shuang Luo, Tianhao Li, Tianyi Tang, Wenbiao Yin, Xingzhang Ren, Xinyu Wang, Xinyu Zhang, Xuancheng Ren, Yang Fan, Yang Su, Yichang Zhang, Yinger Zhang, Yu~Wan, Yuqiong Liu, Zekun Wang, Zeyu Cui, Zhenru Zhang, Zhipeng Zhou, and Zihan Qiu.
\newblock Qwen3 technical report, 2025.
\newblock URL \url{https://arxiv.org/abs/2505.09388}.

\bibitem[Yao et~al.(2023)Yao, Zhao, Yu, Du, Shafran, Narasimhan, and Cao]{yao2022react}
Shunyu Yao, Jeffrey Zhao, Dian Yu, Nan Du, Izhak Shafran, Karthik~R Narasimhan, and Yuan Cao.
\newblock React: Synergizing reasoning and acting in language models.
\newblock In \emph{The Eleventh International Conference on Learning Representations}, 2023.
\newblock URL \url{https://openreview.net/forum?id=WE_vluYUL-X}.

\bibitem[Yen et~al.(2025)Yen, Gao, Hou, Ding, Fleischer, Izsak, Wasserblat, and Chen]{yen2025helmetevaluatelongcontextlanguage}
Howard Yen, Tianyu Gao, Minmin Hou, Ke~Ding, Daniel Fleischer, Peter Izsak, Moshe Wasserblat, and Danqi Chen.
\newblock {HELMET}: How to evaluate long-context models effectively and thoroughly.
\newblock In \emph{The Thirteenth International Conference on Learning Representations}, 2025.
\newblock URL \url{https://openreview.net/forum?id=293V3bJbmE}.

\bibitem[Zhang et~al.(2024{\natexlab{a}})Zhang, Chen, Hu, Xu, Chen, Hao, Han, Thai, Wang, Liu, and Sun]{zhang-etal-2024-bench}
Xinrong Zhang, Yingfa Chen, Shengding Hu, Zihang Xu, Junhao Chen, Moo Hao, Xu~Han, Zhen Thai, Shuo Wang, Zhiyuan Liu, and Maosong Sun.
\newblock $\infty${B}ench: Extending long context evaluation beyond 100{K} tokens.
\newblock In Lun-Wei Ku, Andre Martins, and Vivek Srikumar (eds.), \emph{Proceedings of the 62nd Annual Meeting of the Association for Computational Linguistics (Volume 1: Long Papers)}, pp.\  15262--15277, Bangkok, Thailand, August 2024{\natexlab{a}}. Association for Computational Linguistics.
\newblock URL \url{https://aclanthology.org/2024.acl-long.814}.

\bibitem[Zhang et~al.(2025)Zhang, Li, Long, Zhang, Lin, Yang, Xie, Yang, Liu, Lin, Huang, and Zhou]{qwen3embedding}
Yanzhao Zhang, Mingxin Li, Dingkun Long, Xin Zhang, Huan Lin, Baosong Yang, Pengjun Xie, An~Yang, Dayiheng Liu, Junyang Lin, Fei Huang, and Jingren Zhou.
\newblock Qwen3 embedding: Advancing text embedding and reranking through foundation models, 2025.
\newblock URL \url{https://arxiv.org/abs/2506.05176}.

\bibitem[Zhang et~al.(2024{\natexlab{b}})Zhang, Sun, Chen, Pfister, Zhang, and Arik]{coaneurips20204}
Yusen Zhang, Ruoxi Sun, Yanfei Chen, Tomas Pfister, Rui Zhang, and Sercan~O Arik.
\newblock Chain of agents: Large language models collaborating on long-context tasks.
\newblock In \emph{The Thirty-eighth Annual Conference on Neural Information Processing Systems}, 2024{\natexlab{b}}.
\newblock URL \url{https://openreview.net/forum?id=LuCLf4BJsr}.

\end{thebibliography}
\end{document}